%% file: submde.tex
\documentclass[letterpaper, 10 pt, conference]{ieeeconf}  % Comment this line out
\pdfoutput=1

\IEEEoverridecommandlockouts                              % This command is only
\title{\LARGE \bf
Model-driven engineering approach\\
to design and implementation of robot control system
}

\author{ \parbox{4 in}{\centering Piotr Trojanek\\
         Institute of Control and Computation Engineering\\
         Warsaw University of Technology\\
         Nowowiejska 15/19, 00-665 Warsaw, Poland\\
         {\tt\small P.Trojanek@elka.pw.edu.pl}}
}

\usepackage{listings} % lstlisting fonts looks ugly
\usepackage{flushend}

\input{\PTPhDThesisDir/common.tex}

\renewcommand{\umlFont}[1]{\textsf{\small #1}}
\renewcommand{\fig}[1]{Fig.~\ref{#1}}

\lstdefinestyle{black}{
  basicstyle={\small\fontfamily{pcr}\selectfont},
  commentstyle={\small\fontfamily{ptm}\slshape},
  keywordstyle={\small\fontfamily{pcr}\fontseries{b}\selectfont},
  columns=fullflexible,
  breaklines=true,
  mathescape=true,
  escapechar=\#,
  tabsize=4,
  frame=lines,
  showstringspaces=false,
  captionpos=b,
  float=tbp,
  deletecomment=[l]{--},
  moredelim=[l]{--},
  morekeywords={excludes,notEmpty,forAll,includes,subOrderedSet,indexOf,last,isEmpty,select,includesAll}
}

\usepackage{xcolor}
 \definecolor{theblue} {rgb}{0.02,0.04,0.48}
 \definecolor{thered}  {rgb}{0.65,0.04,0.07}

 \usepackage[
  bookmarks,bookmarksopen,pdfdisplaydoctitle,colorlinks,
  linkcolor=theblue,citecolor=theblue,urlcolor=thered,
  hyperindex=false,hyperfootnotes=false,
  pdfauthor={Piotr Trojanek},
  pdftitle={Model-driven engineering approach to design and implementation of robot control system}
  ]{hyperref}

\begin{document}

\maketitle
\thispagestyle{empty}
\pagestyle{empty}

%%%%%%%%%%%%%%%%%%%%%%%%%%%%%%%%%%%%%%%%%%%%%%%%%%%%%%%%%%%%%%%%%%%%%%%%%%%%%%%%
\begin{abstract}

In this paper we apply a model-driven engineering approach to designing domain-specific solutions for robot control system development.
We present a case study of the complete process, including identification of the domain meta-model, graphical notation definition and source code generation for subsumption architecture -- a well-known example of robot control architecture.
Our goal is to show that both the definition of the robot-control architecture and its supporting tools fits well into the typical workflow of model-driven engineering development.

\end{abstract}

%%%%%%%%%%%%%%%%%%%%%%%%%%%%%%%%%%%%%%%%%%%%%%%%%%%%%%%%%%%%%%%%%%%%%%%%%%%%%%%%
\section{INTRODUCTION}

There is a long history of development of \emph{robot control architectures},
however, there exists no precise definition of this term.
Moreover, their designing is still much more of an art than a science~\cite{RSAaP:08}.
We believe that while one may argue about the ideas behind a particular robot control architecture,
a systematic engineering approach can be applied to the process of designing an architecture itself.

\subsection{Robot control architectures}

The work on robot control architectures began in the late 1960s with the \emph{sense-plan-act} (SPA), a dominant paradigm at that time~\cite{RSAaP:08,brooks:1986_robust}.
This approach, originated from the artificial intelligence community, decomposes a control system into three functional layers dealing
with sensing state of an environment,
planning an optimal action based on symbolic representation of the current state
and finally executing selected action.
Even if robot succeeds extracting a~meaningful representation of a world model from raw sensor data,
it usually takes a long time to plan for correct action.
As a result, the executed action often corresponds to the already outdated sensor readings.
Because of this, the SPA architecture fails to provide the robot with a robust operation in the real world, noisy and unpredictable environments.

In the mid-1980s limitations of the SPA architecture were recognized and the robotics community turned towards
more reactive, behavioural approaches to robot control.
Probably the most wide-known example is the \emph{subsumption architecture} of Brooks~\cite{brooks:1986_robust}.
A subsumption architecture is built from a~set of small processors (referred to as modules) which sends messages to each other.
These modules make up \emph{layers} of control corresponding to different levels of competence, e.g.,
dealing with obstacle avoidance, wandering, exploration, map building and navigation.
The subsumption architecture proved to be successful in operating within unpredictable, noisy environments.
However, it was difficult to assign a~long-term, complex task to a robot executing just a set of relatively low-level behaviours.

Limitations of behaviour-based controllers led to the development of layered robot architectures~\cite{gat1998three}.
These architectures combine real-time control, action sequencing and planning
as bottom, middle and top layers, respectively.
The challenge of developing a layered control system consists both of
representing and solving problems at the individual layers,
as well as integrating individual levels of control.
Because of different requirements and constraints,
a typical approach is to use dedicated domain-specific languages at individual layers~\cite{RSAaP:08}.
%Examples of such a languages include ALFA~\cite{gat1991alfa} and Rex~\cite{kaelbling1987rex} for reactive control,
%ESL~\cite{gat1997esl} and TDL~\cite{Simm:98} for action execution
%and PRS~\cite{ingrand1996prs} and PLEXIL~\cite{verma2005plan} for planning.
While portable middleware and APIs allow for software reuse between different architectures~\cite{Brugali:2009},
in most cases it is still difficult to separate architectural concepts from their implementation.

\subsection{Model-driven engineering}

Model-driven engineering (MDE) technologies combine~\cite{schmidt2006model}:
\begin{itemize}
\item \emph{Domain-specific modeling languages} (DSMLs) whose type systems 
formalize the application structure, behaviour, 
and requirements within particular domains.
\item Transformation engines and generators that analyze 
certain aspects of models and then synthesize various
types of artifacts, such as source code.
\end{itemize}

DSMLs are typically specified with meta-models,
which have significant advantages over other techniques,
such as BNF grammars or UML profiles~\cite{kleppe2009software}.
The choice of a meta-modeling technology depends mainly on the availability of tools supporting the development of a complete solution.
\mbox{MetaEdit+} was among the first commercially available domain-specific development environments~\cite{kelly2008domain}.
Increasing market interest has been recognized by Microsoft, thus the release of Microsoft Domain-Specific Language Tools in 2005~\cite{msdsl2007}.
The Eclipse Modeling Framework~\cite{emf2009,gronback2009eclipse} is the leading open-source competitor in the field.
It provides advanced support for both graphical and textual notation of modeling languages,
as well as
validation of model constraints,
model-to-model and model-to-text transformations.
Individual tools of the framework follow the Object Management Group (OMG) standards~\cite{MOF,OCL,MOFM2T:08}.

This paper is structured as follows.
Section \ref{emf-overview} presents an overview of the Eclipse Modeling Framework
-- one of the most popular model-driven engineering technologies available.
Section \ref{domain-specific-solution} presents application of the meta-modeling approach to capturing concepts of an exemplary robot control architecture.
A complete environment for development of a robot control system,
including generation of application code skeleton, is presented.
Section \ref{related-work} compares this work with existing approaches.
Section \ref{conclusions} presents conclusions and remarks.

%Everything that can be thought at all can be thought clearly. Everything that can be put into words can be put clearly.

\section{THE ECLIPSE MODELING FRAMEWORK (EMF) PROJECT OVERVIEW}
\label{emf-overview}

The EMF project is a modeling framework and code generation facility for building tools and other applications based on a structured data model~\cite{emf2009}.
The accompanying projects enable the development of a complete modeling solution, including
textual and graphical editors,
validation of model constraints,
model-to-model and model-to-text transformations~\cite{gronback2009eclipse}.

Ecore, the underlying data model of EMF, is aligned on the Meta-Object Facility OMG standard~\cite{MOF}.
The Ecore model includes elements known from object-oriented programming, such
as classes, attributes and references (i.e., composition, inheritance and associations)~\cite{emf2009}.
The cardinality of attributes and references is expressed with lower and upper bounds.
Complex model restrictions are specified by the use of Object Constraint Language (OCL)
expressions~\cite{OCL,warmer2003object}.
Constraint statements are defined in the contexts of an individual classes and specify invariants related to their attributes and relationships.

The EMF models are typically presented with graphical notation, which is based on UML class diagram notation~\cite{umlSuper23}.
However, in opposition to the models created in UML, the models created with meta-modeling solutions are formal and precise~\cite{kelly2008domain}.

The EMF is accompanied by several satellite projects, which support different steps of a typical model-driven engineering workflow.
These include tools for the development of graphical and textual notations,
languages for model-to-model and model-to-text transformations,
model validation and constraint checking~\cite{gronback2009eclipse}.
It is important to note, that these tools can be used interchangeably or in addition to those
which have been applied in the following use case.

%\begin{table}[htbp]
%\center
%\begin{tabular}{ | l | c | c| c | }
%\hline
%Type of relationship & Diagram symbol \\
%\hline
%\hline
%  inheritance & \tikz{\draw[inherit] (0,0) -- (5em,0);} \\
%  \hline
%  composition & \tikz{\draw[compose] (0,0) -- (5em,0);} \\
%  \hline
%  reference & \tikz{\draw[refer] (0,0) -- (5em,0);} \\
%\hline
%\end{tabular}
%\caption{Types of relationships and their
%%corresponding
%diagram symbols}
%\label{class-relation-symbols}
%\end{table}
%
%\begin{figure}[htb]
%\center
%\resizebox{\columnwidth}{!}{
%\input{\PTPhDThesisDir/ecore-m3.tikz}
%}
%\caption{Ecore diagram of the used meta-metamodel subset}
%\label{ecore-m3}
%\end{figure}

\section{DOMAIN-SPECIFIC SOLUTION}
\label{domain-specific-solution}

The subsumption architecture is one of the most well-known robot control architectures.
While its limitations have been already identified~\cite{gat1998three},
it is still introduced at many preliminary robotics courses.
Subsumption systems were originally intended for hardware implementation as a network of cooperating processors.
Later, the same set of concepts was implemented as the \textit{Behaviour language}~\cite{brooks1990behavior} --
an extension to Common Lisp that enabled both
the execution of the program on the host machine
and the compilation into microcontroller assembly code, which was executed on-board the robot.

In the original paper by Brooks, the architectural concepts were introduced informally and without detailed definition of their semantics~\cite{brooks1990behavior}.
This approach sufficed to attract interest in the new approach to robot programming,
however, it should not be regarded as the approved way to define robot control architectures.
The following section describes domain-specific solution for the development of a subsumption-based robot control system.
Individual subsections follow format of presentation found in~\cite{kelly2008domain}.
Our goal is to show that both the definition of the robot-control architecture and its supporting tools fits well into the typical workflow of model-driven engineering development.

\subsection{Introduction and objectives}

The subsumption architecture differs fundamentally from the typical \emph{sense-plan-act} approach,
which was initially the predominant architecture.
One of the main differences is that it breaks a typical sequential data flow that simply links sensors to actuators.
Instead, a subsumption-based robot control system consists of a set of asynchronous processors (modules)
interconnected with wires, which enable both data and control flow.

Internal operation of a processor is described by an augmented finite state machine (AFSM)~\cite{brooks:1986_robust}.
The processor is capable of:
(1)~sending a new data message with an output wire,
(2)~modifying its internal memory variables,
(3)~conditional dispatching based on predicates on it's internal variables and values of the input buffers,
as well as (4)~event dispatching based on monitoring input wires for arrival of a new message with an optional timeout.
The latest data from an input wire are held in a module's input buffer.
However, the exact semantics of the AFSMs is not specified formally.

The core architectural concepts are related not to the individual processors, but to their operation as a network.
The processor's operation can be easily described using any of the existing general-purpose programming languages
extended with library calls for inter-module communication.
In fact, the \textit{Behaviour language} did not specify the AFSMs directly, but rather with sets of real-time rules~\cite{brooks1990behavior}.
However, in this paper we focus only on the architecture and not on the patterns for building the AFSMs.

\subsubsection{Target Platform and Environment}

While it is possible to implement a robot control system directly in hardware (e.g., using gate arrays or analog circuits),
most of the robots are equipped with on-board computers.
On-board processor systems range from a single microcontroller (e.g., Lego Mindstorms NXT)
to a cluster of PC-s (e.g., autonomous vehicles in the DARPA Grand Challenge competition~\cite{Thrun:2006Stanley}).

The ultimate goal of a robot control system architecture is to provide developers with a skeleton,
which can be just filled with data types definitions and implementation of control algorithms.
Because of the multitude of existing operating systems and middleware available,
there is no a~single platform for the deployment of a control system.
Instead, code portability is the most valuable feature and one of the main concerns.
%While it is possible to rely the control system on one of the well supported operating system API (Application Programming Interface)
To increase the portability of the control software,
the generated application software should not be specific to any particular operating system,
but should be layered on top of a well-defined programming interface.
In addition, a~well-chosen interface (target environment) can significantly decrease complexity of model-to-code transformations.
An example of such an interface is provided by the POSIX standard,
which defines both the API and the set of tools, required to build an executable binary~\cite{POSIX-1.2008}.
However, the environment defined by the POSIX is of a relatively low level
and at the same time requires a significant support on the side of the underlying operating system
(e.g., dynamically created threads and synchronization primitives).

In our approach we have decided to use the Ada programming language as a target of code generation~\cite{taft2006ada}.
Ada is a~general-purpose programming language, originally targeted at embedded and real-time systems;
it has been already successfully applied to robot control~\cite{Gin:82,steele1994ada}.
Our choice is motivated mainly by the wide set of features supported
at the language level, including advanced multi-tasking and distributed systems.

In particular, two extremes of robot platforms are extensively supported by Ada, namely,
deeply embedded systems with limited resources and large, heterogeneous distributed environment. % the Ada programming language.
The former case is addressed by the \emph{Ravenscar profile}, which is a subset of the language restricted to met
the requirements for determinism and schedulability analysis, as well as being suitable for mapping to small runtime
systems of resource-limited platforms~\cite{burns2004guide}.
The latter case is addressed by the \emph{Distributed Systems Annex}, which enables the execution of a multi-tasking Ada program
on a set of distributed processing nodes~\cite{vergnaud2004polyorb}.
Both cases require annotations of Ada code with standard-defined pragmas (compiler directives).
The pragmas related to the Ravenscar profile are compatible with those related to the Distributed Systems Annex.
Thus the same source code can be used to build both executables for distributed system,
as well as an executable for a resource-limited embedded board.
This flexibility and portability was our main motivation for using Ada,
however, other targets for the code generation are possible as well.

\subsubsection{Objectives}

The objective of this modeling solution is to ease the development of the subsumption based robot control systems.
This goal is achieved by allowing the designer to specify the structure of the control system in an intuitive graphical notation
and then to automatically generate a code skeleton of the final application.
The code skeleton has to be filled with routines (portions of code) with
implementation of the data processing algorithms and interface with the robot's sensors and actuators.

A model-driven approach allows to generate multiple artifacts from a single source model.
Thus it is possible to provide the designer also with additional model-to-code transformations,
e.g., generation of a documentation templates or unit tests.

\subsection{Development process}

Development of a domain-specific solution starts with identification of the modeling concepts.
In the case of robot control architectures the main references are publications with details of a given architecture.
Evaluation reports and source code of existing applications can be also valuable sources of information,
but these are not always publicly available.
Moreover, authors typically point out the core architectural concepts and their attributes,
but concept relationships and their constraints are often introduced only informally.
Thus, the main challenge usually is to find relationships of the architectural concepts, their type and cardinality.

The naming used within a domain-specific solution usually follows the naming of the architecture's authors.
However, some of unnamed relationships (and opposite relationships in particular)
are often required for navigation in the model transformations.
In these cases, the developer of a domain-specific solution can use arbitrary names,
since they will not be directly visible to end users.

For the robot control architectures, there are typically no more than several core concepts and relationships to model.
Authors tend to intentionally avoid architectural complexity,
which would result in structures that are difficult to analyze and understand by end users.
In the case of subsumption architecture the main effort in creating a domain-specific solution
was to develop a graphical notation editor corresponding to the original notation
and model-to-code transformation, rather than identifying the meta-model of the domain.

\subsection{Modeling language}

The domain-modeling language for subsumption-based robot control system collects
concepts and their relationships from the original publications~\cite{brooks:1986_robust,brooks1990behavior}.
The benefit of using meta-modeling for language specification is a much clearer presentation of its structure
compared to informal description in natural language only.

\begin{figure*}[thpb]
\centering
\resizebox{.83\textwidth}{!}{
\input{subsumption-meta.tikz}
}
\caption{Subsumption architecture meta-model}
\label{subsumption:meta-model}
\end{figure*}
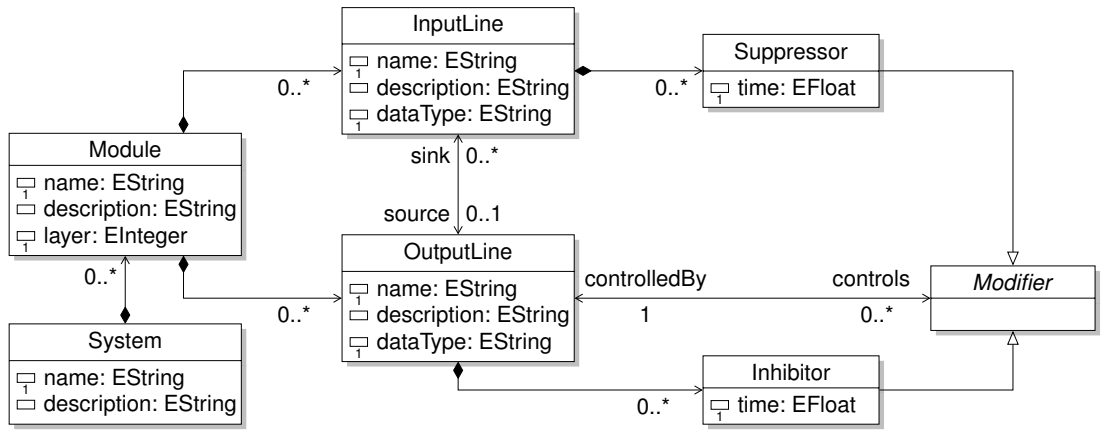

\subsubsection{Modeling concepts}

The root element of the subsumption-based robot control system modeling language definition is \umlFont{System} itself,
as it contains instances of all other elements of the language (\fig{subsumption:meta-model}).
The mandatory \umlFont{name} attribute is used to identify an instance, while the optional \umlFont{description} is included for documentation purpose only.
These general attributes will be common also to other concepts of the domain;
they are used to generate names and comments in code skeletons and documentation.

The \umlFont{System} is composed of \umlFont{Modules}, each corresponding to a single activity within a control system.
The mandatory \umlFont{layer} attribute specifies to which \emph{level of competence} a~particular module belongs to~\cite{brooks:1986_robust}.
The purpose of this attribute is to facilitate testing of the control system by separately enabling or disabling individual layers in runtime.
%The \umlFont{period} is related to event dispatch activity of the finite-state machine running within a single processor.
%Semantics of this attribute is as follows.
%A positive value corresponds to the cycle time of module's internal periodic activity.
%If the value is set to zero, then the module's internal activity will be triggered only by a new data arriving on any of the module's input wires.
%A negative value specify timeout on waiting for a new data arrival.
Each \umlFont{Module} contains both \umlFont{InputLine}s and \umlFont{OutputLine}s,
which can receive and send data from/to other modules, respectively.
\umlFont{DataType} attribute specifies the type of messages transmitted with the individual lines.
It is required that the data type of an input line is convertible to a data type of a corresponding output line.
The strong typing feature of the Ada type system enables checking this requirement at compile time, thus enforces matching of modules' interfaces.

A connection wire between \umlFont{InputLine} and \umlFont{OutputLine} is modeled with the \umlFont{source} and \umlFont{sink} relationships.
The upper bound on the cardinality of the \umlFont{source} relationship equals to one --
each \umlFont{InputLine} may be wired to at most one \umlFont{OutputLine}.
On the other hand, there is no upper bound on the cardinality of the \umlFont{sink} relationship --
each \umlFont{OutputLine} can be wired to several \umlFont{InputLine}s.

The key feature of the subsumption architecture is the ability to \emph{suppress}
data incoming into one module's input by data outgoing from another module's output
and similarly to \emph{inhibit} one module's output data with another's.
These features are modeled with \umlFont{Suppressor} and \umlFont{Inhibitor} concepts,
which intercept transmission on the wires incoming to \umlFont{InputLine} and outgoing from \umlFont{OutputLine}, respectively.
The interception is done by messages outgoing with \umlFont{OutputLine} specified with the \umlFont{controlledBy} relationship.
This relationship is common for both \umlFont{Suppressor} and \umlFont{Inihibitor},
thus it has been assigned to a common abstract \umlFont{\textit{Modifier}} concept.
The \umlFont{time} attributes specify for how long after the arrival of a message on the control line the data transmission remains intercepted.
They have not been assigned to the \umlFont{Modifier} concept, since they are distinct properties of the \umlFont{Suppressor} and \umlFont{Inihibitor}.

\subsubsection{Modeling rules}

In addition to concepts, their attributes and relationships, it is often necessary to include more detailed restrictions on the domain meta-model.
For the ECore they can be described using the Object Constraint Language (OCL) expressions~\cite{OCL,warmer2003object}.
OCL is a declarative language that enables one to precisely specify constraints on the model, e.g., values of attributes and scope of relationships.

%Additional, more complex model restrictions can be specified using Object Constraint Language (OCL)
%expressions~\cite{OCL,warmer2003object}.
%Constraint statements are defined in the context of an individual class and specify invariants related to their attributes and relationships.

In the case of the subsumption control architecture meta-model, OCL expressions are used to
limit the scope of the interception relationships (i.e., suppressors and inhibitors)
and the allowed range of the numerical attributes (listing \ref{SubsumptionOCL}).
OCL expressions typically consist of the \emph{context} keyword followed by a class name
and \emph{inv:} keyword followed by class invariant definitions.
Dot notation, similarly as used in object-oriented programming,
enables the navigation within a~model to specify a constraint on a related concept.
The \emph{self} keyword is used to address the context's attributes explicitly.

\begin{lstlisting}[language=OCL, style=black,
	caption=OCL invariants imposed on the domain's meta-model,
	morekeywords={excludes,notEmpty},
	texcl=true,
	label=SubsumptionOCL
]
context Module
-- $\textrm{\fontfamily{ptm}\slshape it is only possible to intercept data transmission}$
-- $\textrm{\fontfamily{ptm}\slshape on the same layer and below}$
inv: outputs.controls.oclAsType(InputLine).
	Module.layer <= self.layer
inv: outputs.controls.oclAsType(OutputLine).
	Module.layer <= self.layer

-- $\textrm{\fontfamily{ptm}\slshape transmission's interception time is always positive}$
context Inhibitor
inv: time > 0

context Suppressor
inv: time > 0
\end{lstlisting}

% Model checking reports

\subsubsection{Modeling notation}

Notation is one of the most important aspects of the domain-specific modeling solution from the end user's point view.
It is worth to note, that it is possible to have multiple notations for a single modeling language.
Different notations may be useful for users depending on their experience level
or their roles in the development process (e.g., project managers and developers).
In particular, in the inter-disciplinary domains such as robotics, it may be beneficial to
discuss the high-level models using an intuitive, graphical notation (possibly with some details hidden from the view).
%Experienced software developers will probably prefer textual notation, similar to programming languages
%they are using on their everyday work.

Notation does not influence in any way the expressive power of a domain-modeling language.
However, non-intuitive diagrams (in the case of graphical notation) or obscured syntax (in the case of textual notation)
can easily discourage potential users from adopting the solution.
The EMF project supports development of both graphical~\cite{gronback2009eclipse} and textual~\cite{efftinge2006oaw} notations for modeling languages.
A prototype graphical notation for a~given domain meta-model can be built with easy to use wizards
and then customized, e.g, with domain-specific symbols.
Prototype textual notation can be also generated automatically~\cite{HUTN} and then customized.

In general, it is much easier to develop both graphical and textual notation from scratch,
than adapt automatically generated prototypes to match pre-existing conventions.
However, in this paper we follow the latter approach to show
that model-based engineering can be also applied to legacy domains.
The graphical notation we adapt to is taken directly from the Brooks original paper~\cite{brooks:1986_robust}.
While this notation is well-known among the robotics community, to our best knowledge,
there exists no graphical environment, which would allow for development of the subsumption-based robot control systems.

\begin{figure*}[tb]
\centering
\includegraphics[width=\textwidth]{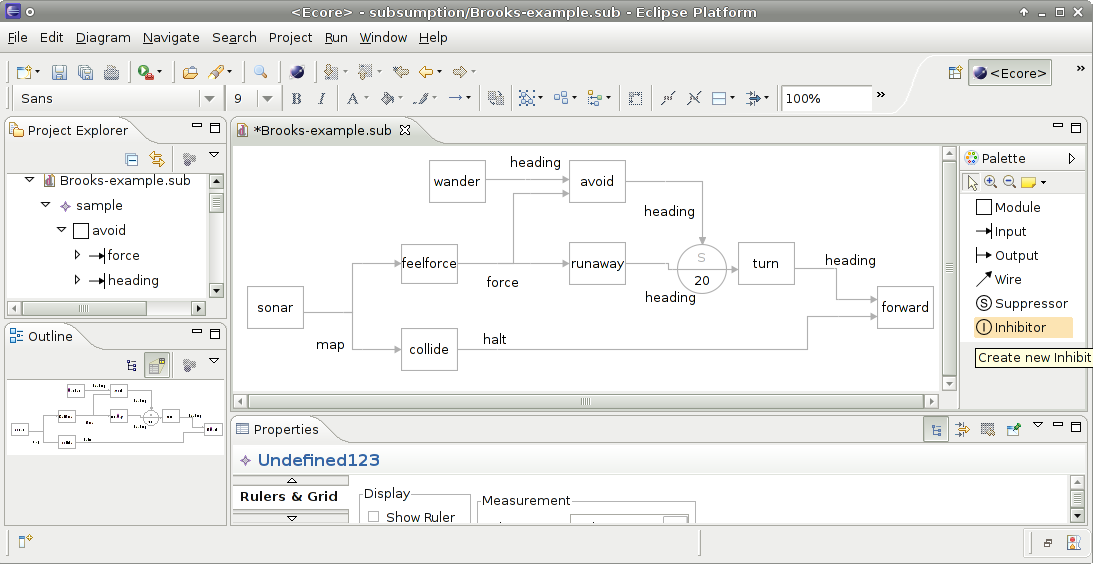}
\caption{Graphical notation of the exemplary subsumption-based control system}
\label{brooks-example}
\end{figure*}

The graphical modeling notation of a subsumption-based control system consists of a single diagram (\fig{brooks-example}).
The root element of EMF diagram is the canvas, which correspond to the root element of the domain meta-model.
A single canvas contains a number of figure definitions,
which are then referenced by logical elements of the diagram,
i.e., nodes, connections, compartments and labels.
Figures are typically defined with a set of predefined shapes (e.g., rectangles, ellipses, polylines)
and their attributes (e.g., width, color, size).
However, it is also possible to create custom figures using Java and the underlying graphics toolkit.
Logical elements of the diagram are then mapped to the domain meta-model.
Nodes are typically mapped to domain concepts,
connections to domain concepts' relationships
and labels to concepts' attributes.
Compartments are designed as containers for other diagram elements
and typically are mapped to the containment relationships of the domain meta-model.

\begin{figure}[tb]
\centering
\includegraphics[width=\columnwidth]{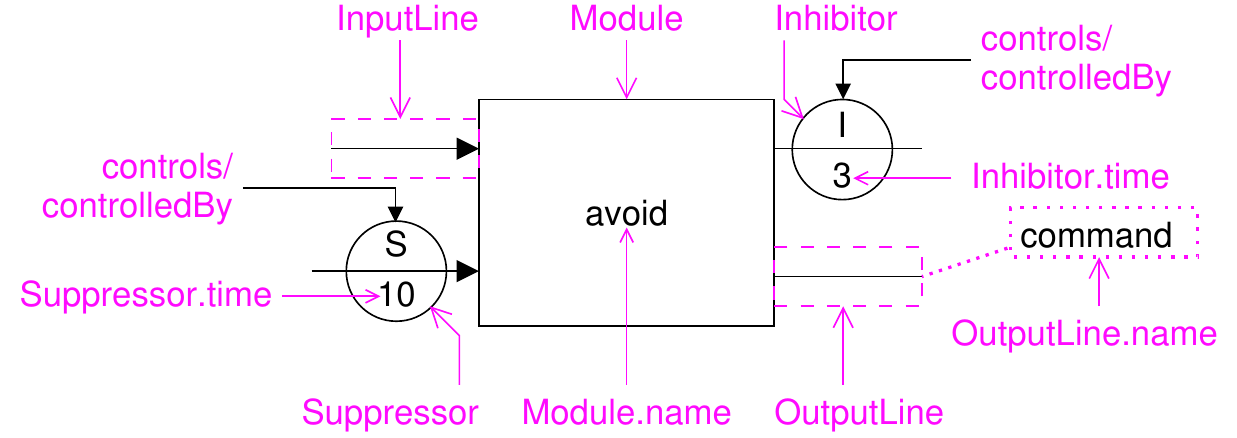}
\caption{Graphical notation of the modeling concepts and their attributes}
\label{module-layout}
\end{figure}

In the subsumption-based control system notation the \umlFont{System} (root element)
is represented with a white background canvas.
Individual \umlFont{Module}s are represented by rectangles with centered label containing the module's \umlFont{name} attribute (\fig{module-layout}).
Both \umlFont{InputLine} and \umlFont{OutputLine} concepts are represented by compartments of rectangular shape
with line connecting their left and right sides
(in the case of \umlFont{InputLine} there is also an arrow head at the right end of the line).
\umlFont{InputLine}s can be placed only on the left border of the module's figure,
while \umlFont{OutputLine}s only on its right border.
The \umlFont{name} attribute of the \umlFont{OutputLine} concept is represented by floating label,
similar to the original Brooks notation.
Both the \umlFont{Suppressor} and \umlFont{Inhibitor} concepts are represented by circles,
respectively with the \textsf{S} and \textsf{I} characters and labels corresponding to their \umlFont{time} attribute.
The \umlFont{sink/source} relationships are represented by lines connecting the corresponding \umlFont{InputLine} and \umlFont{OutputLine} diagram elements.
The \umlFont{controls/controlledBy} relationship is represented by arrows from the \umlFont{OutputLine} to one of the \umlFont{Suppressor} and \umlFont{Inhibitor} diagram elements.
Relationship connections can be anchored only at specified locations of the diagram elements
(top in the case of \umlFont{Suppressor} and \umlFont{Inhibitor} figures,
while left and right in the case of \umlFont{InputLine} and \umlFont{OutputLine}, respectively).
A~palette with tools mapped to individual domain concepts and relationships is also included
and can be easily customized with icons corresponding to the domain elements.

The customized graphical notation solution built with the EMF project consists also of many generic features,
which otherwise had to be implemented manually.
These include outline view of the diagram and structural view of the model,
diagram grid, snapping connections to anchor points, zooming,
automatic layout and alignment of diagram elements, printing, exporting to bitmap and vector image formats (\fig{brooks-example}).
Diagrams and models can be stored using the XMI (XML Metadata Interchange) file format,
thus an interoperation with tools from outside of the EMF project is also possible.

\subsection{Modeling subsumption control system}

The presented domain-specific modeling solution has been tested by developing an exemplary control system for an autonomous mobile robot.
Both the application scenario as well as design of modules and their interconnections has been reused from the original Brooks paper~\cite{brooks:1986_robust}.
Our main motivation was to keep the reader's focus on the approach itself, rather than on the details of the implemented application scenario,
robot design or control algorithms.

One of the main drawbacks of the subsumption architecture is that a controller is not task-oriented (and non-taskable).
Rather, the behaviour of the robot emerges from interactions of controller's modules.
Thus, the goal of the mobile robot is simply to autonomously operate in an unstructured environment,
including obstacle avoidance and possibly exploration of interesting locations.
Similar to the original paper, experiments have been performed both on simulated and real robots.

It should be stressed that the goal of this paper is not the evaluation of the subsumption architecture or the particular instance of the controller.
The focus of the paper is rather on evaluation of model-driven engineering approach to designing robot control architectures.

\subsubsection{Example model}

The exemplary model consists of two layers of the original control system presented by Brooks,
i.e., avoiding contact with objects and wandering aimlessly without hitting obstacles~\cite{brooks:1986_robust}.
Exploration (the third layer) requires more advanced sensors (in the original paper a tilt head with stereo camera has been used)
and data processing algorithms (image feature extraction, path planning).

Individual modules are responsible for:
filtering and processing sensor range data into robot centered map of obstacles (\textit{sonar}),
detecting collisions (\textit{collide}),
computing repulsive forces from obstacles (\textit{feelforce}) and monitoring their sum (\textit{runaway}),
generating a new heading for the robot (\textit{wander}) and combining it with the data about obstacles (\textit{avoid})
(\fig{brooks-example}).
Finally, \textit{turn} and \textit{forward} modules interact with the motors.
A single suppressor intercepts heading generated by the \textit{runaway} module with that generated by the \textit{avoid}.

\subsubsection{Use scenario}

Modeling an instance of subsumption-based control system begins with the definition of individual modules
together with their output and input lines.
Afterwards, the output and input lines are connected by wires.
Suppressor and inhibitors are created, assigned activation time and their control lines as required.
Names and descriptions in the model are also required, as they are used to generate the source code.
Individual OCL rules are enforced during model development.

When the model definition is completed, then it is possible to automatically generate skeleton of the robot control system application.
Individual modules are translated into separate Ada packages, thus the structure of the source code follows the structure of the model
and the code corresponding to each module can be compiled separately.
A user is required to provide (1) specification of the data types assigned to input and output lines
and (2) bodies of procedures that implement AFSMs executed within individual modules.
Implementation typically begins from modules responsible for getting sensor data
and can be incrementally tested, even if another modules are not yet completed.

\subsection{Generator for Ada}

Generator for the robot control system has been implemented in Acceleo, an Eclipse implementation of the OMG Model to Text Transformation language~\cite{MOFM2T:08}.
Transformation consists of navigating an instance of the control system model, starting from the root \umlFont{System}
and applying a template of Ada code to the concepts of the domain meta-model.
Ada code templates have been extracted from a hand-written implementation of a simple subsumption-based control system.
The generator produces both the source code and a project configuration file for the Ada IDE editor.

\subsubsection{Generator structure}

The generator structure corresponds to the structure of the initial, hand-written Ada source code.
Individual templates generate packages with the implementation of the module's interface (procedures to send data with input lines),
input buffers (where new data values are stored until received by AFSM),
and main task (which executes AFSM).
A separate template generates specifications of procedures, which implements internal operations (AFSMs) of the modules.

\subsubsection{Generator in action}

%710 - mtl.
%1009 - SLOC (Ada).
%1968 - wszystko.
The complete generator consists of 9 files with a total of about 700 lines of Acceleo's template code.
For an exemplary controller with 8 modules, about 1800 source lines of Ada code (excluding white space and comments) were generated.
This code is responsible for the initialization of tasking constructs, input line buffers and connection wires.
The internal operation of modules (AFSMs) has been manually implemented with about 200 lines of code,
which is only about 10\% of the complete application.
It should be noted that all aspects of tasking and concurrency (which are relatively more difficult to develop and debug)
are handled by the automatically generated code.
Thus, a developer can focus on data processing algorithms only and not on implementation of the architectural concepts.

\subsection{Framework support}

The generated source code purposely does not rely on any software framework
other than standard features provided by the Ravenscar profile subset of Ada,
i.e., tasks and protected types~\cite{burns2004guide}.
This allows to build the control application for resource limited and embedded platforms,
which lack features like terminal console or filesystem.
Generated code automatically benefits from using multi-core processors, since every module is executed within a separate Ada task.
It is also possible to compile the code into a distributed application~\cite{vergnaud2004polyorb}.
In this case individual modules can be transparently executed on separate network nodes
(e.g., computationally expensive data processing can be off-loaded).

It is possible to provide alternative generators for platforms with more features available
(e.g., data logging, interactive debug console, runtime configuration files).
In such case a separate software framework should be provided to avoid unnecessary complexity of a generator's templates.

\subsection{Main results}

The exemplary control system has been tested with both simulated and real-world robots.
Because of limitations of the Lego Mindstorms NXT platform only three range finders were used
(in the original Brooks research a ring of twelve sonar sensors has been used).

\begin{figure}[tb]
\centering
\includegraphics[width=\columnwidth]{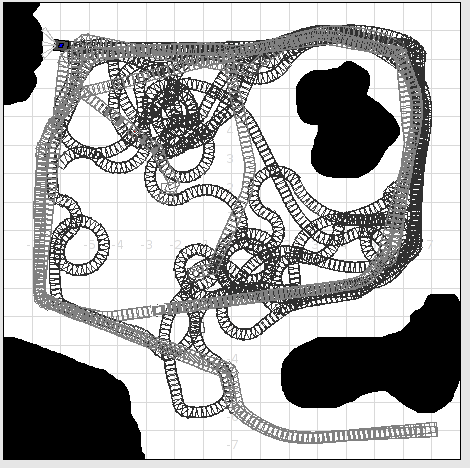}
\caption{Paths of the subsumption controlled robot in simulated environment
with only the first one (light green) and both layers enabled (dark green)
}
\label{stage-path}
\end{figure}

For the simulation the \emph{Player/Stage} software suite has been used~\cite{vaughan2008massively}
together with client library bindings for Ada. %\cite{MosteoPlayerAda}.
In the simulated environment two versions of controller have been tested,
with the first competence layer disabled and enabled (\fig{stage-path}).
Performance of the controller strongly depends on the parameters of individual modules
(i.e., collision distance, repulsive force settings).
However, the layer with \textit{wander} behaviour clearly cause the robot to operate more robustly and explore a larger area.

For real-world experiments the Lego Mindstorms NXT platform has been used,
which provides only 64kB of RAM and 256kB of flash memory.
An executable image, which consists of Ada Ravenscar runtime, NXT device drivers,
internal code of the subsumption modules and the inter-module data buffers,
is only about 50kB in size.
The Ravenscar profile does not put any restrictions
on the \emph{sequential} (i.e., non-tasking) features of Ada and the architectural footprint of the generated application is minimal.

%Because of limited resources only the first layer has been implemented,
%since the \emph{Ravenscar} runtime together with platform drivers and tasks of multiple modules consumes almost all of available memory.
%Moreover, the current release of the GNAT Ada Mindstorms contains bug, which does not allow to access more than one sonar sensor at a time.
%Two reflected-light sensors has been used as a workaround, however, they allowed only to detects obstacles with a relatively short distance.
%Despite these limitation a successful trials with the NXT robot has been executed.
%However, further investigation and optimization of the platform settings is required in order to evaluate more complex applications.

%\begin{figure}[htb]
%\center
%%\includegraphics[width=\columnwidth]{robot-nxt-makro}
%\includegraphics[width=\columnwidth]{nxt-3sonars}
%%
%\caption{Lego Mindstorms NXT robot with sonar range finder and two light sensors, as used in experiments}
%\label{robot-nxt-makro}
%\end{figure}

%%%%%%%%%%%%%%%%%%%%%%%%%%%%%%%%%%%%%%%%%%%%%%%%%%%%%%%%%%%%%%%%%%%%%%%%%%%%%%%%
\section{RELATED WORK}
\label{related-work}

The subsumption architecture has been originally implemented as a set of extensions to Common Lisp.
These extensions allowed both for generation of assembly code for a target hardware as well as a native execution, e.g., simulation~\cite{brooks1990behavior}.
Lisp, which was the language of choice among the artificial intelligence researches at that time,
is particularly well suited to the implementation of a domain-specific solutions due to its extensibility features.
Unfortunately, it does not leave any choice regarding the syntax convention.
It is also difficult to distinguish between the domain-specific (architectural) and native Lisp constructs within a single notation.
%However, Lisp has been used for implementation of a 

In~\cite{fleury|iros97}, the authors present a solution to specify robot control modules and generate code skeletons.
However, implementation of a complete toolset
(i.e., parser for textual notation, generator templates and engine, syntax description files for common text editors)
from scratch is time consuming, thus this approach is not widely accepted.
In~\cite{arias2010orccad} the EMF project tools are used to develop a modeling solution,
including domain meta-model, graphical notation editor and generators for multiple artifacts (e.g., code and documentation).
In~\cite{steck2010towards}, the authors use UML profiles to apply interaction patterns within a component-based robot control framework.

These works, however, target \emph{horizontal} domains,
i.e., they do not raise abstraction level significantly nor present the problem in a more readable fashion.
The real benefit and productivity improvement of domain-specific modeling is when applied to \emph{vertical} domains,
i.e., when an intuitive graphical notation and high-level concepts allows the developer to focus on the design, not on the implementation.

\section{CONCLUSIONS}
\label{conclusions}

In this paper we have presented a case study of a complete model-driven engineering process applied to the development of a domain-specific solution for an exemplary robot control architecture%
\footnote{A complete source code, together with documentation of the presented modeling solution are available for download at \url{http://github.com/ptroja/subsumption}}.
The model-driven engineering and DSMLs are relatively new concepts in the robotics research.
For a long time the robotics community has been developing custom solutions
that substitute the steps of a typical model-driven workflow.

Most of the existing approaches to robot control can be summarized as providing patterns
for dealing with the underlying complexity and describing control systems more concisely.
DSMLs allows to capture these patterns, their relationships, meanings and symbols as
a language vocabulary, grammar, semantics and notation, respectively~\cite{kleppe2009software}.
Software language engineering provides methods and tools, which can be applied
to the development of these languages.
In particular, it enables
the separation of concerns of the typical development workflow (e.g., concrete syntax, model validation, code generation)
and
providing alternative implementations of the individual components (e.g., multiple notations and generators dedicated for specific platforms).

DSMLs enables one to create new semantic levels that describe certain abstractions more concisely.
It is then possible to develop transformations (e.g., to check for validity) and generators (e.g., to translate into source code)
for models created with these languages.
In addition, semantics of DSMLs can be precisely specified with model-to-model transformation by providing
mapping into domain with already defined semantics (however, this has not been explored in this paper).
Finally, all these tools can be combined and delivered as a single Integrated Development Environment (IDE).

We believe, that model-driven engineering provides the necessary framework
that enables the development of methods and tools for robot control and programming in a much more disciplined and efficient manner than before.
Dedicated tools for all the steps involved (i.e., definition of domain meta-model, multiple notations, model-to-model and model-to-text transformations)
promise that the final solution can be delivered with much less effort.

%An interesting opportunity of the (which has been not investigated in this paper) is to define
%At this moment it is not possible to easily implement filtering sections of the diagram (e.g. hiding individual layers).

%%%%%%%%%%%%%%%%%%%%%%%%%%%%%%%%%%%%%%%%%%%%%%%%%%%%%%%%%%%%%%%%%%%%%%%%%%%%%%%%
%\section{ACKNOWLEDGMENTS}
%
%The authors gratefully acknowledge the contribution of National Research Organization and reviewers' comments.

%%%%%%%%%%%%%%%%%%%%%%%%%%%%%%%%%%%%%%%%%%%%%%%%%%%%%%%%%%%%%%%%%%%%%%%%%%%%%%%%

%References are important to the reader; therefore, each citation must be complete and correct. If at all possible, references should be commonly available publications.

%\begin{thebibliography}{99}
%
%\bibitem{c1}
%J.G.F. Francis, The QR Transformation I, {\it Comput. J.}, vol. 4, 1961, pp 265-271.
%
%\bibitem{c2}
%H. Kwakernaak and R. Sivan, {\it Modern Signals and Systems}, Prentice Hall, Englewood Cliffs, NJ; 1991.
%
%\bibitem{c3}
%D. Boley and R. Maier, "A Parallel QR Algorithm for the Non-Symmetric Eigenvalue Algorithm", {\it in Third SIAM Conference on Applied Linear Algebra}, Madison, WI, 1988, pp. A20.
%
%\end{thebibliography}

%\bibliographystyle{plain}
\bstctlcite{IEEEexample:BSTcontrol}
\bibliographystyle{IEEEtran}
\bibliography{../../common_utf8/robot.bib}

\end{document}

%% file: common.tex
\usepackage{tikz}
\usetikzlibrary{arrows,positioning,fit,calc,shapes.multipart,shadows,decorations.pathreplacing,shapes,decorations.shapes,decorations.markings}

\makeatletter
\pgfarrowsdeclare{blocktip}{blocktip}
{
  \pgfmathsetlength{\pgfutil@tempdima}{.25\pgflinewidth+.25*\pgfinnerlinewidth}%
  \pgfmathsetlength{\pgfutil@tempdimb}{.5\pgflinewidth-.5*\pgfinnerlinewidth}%
  \pgfarrowsrightextend{2\pgfutil@tempdima+.5\pgfutil@tempdimb}
  \pgfarrowsleftextend{1.3\pgfutil@tempdima+.5\pgfutil@tempdimb}
}
{
  \pgfmathsetlength{\pgfutil@tempdima}{.25\pgflinewidth+.25*\pgfinnerlinewidth}%
  \pgfmathsetlength{\pgfutil@tempdimb}{.5\pgflinewidth-.5*\pgfinnerlinewidth}%
  \pgfsetlinewidth{\pgfutil@tempdimb}
  \pgfsetdash{}{+0pt}
  \pgfsetroundcap
  \pgfsetroundjoin
  \pgfpathmoveto{\pgfpoint{0}{\pgfutil@tempdima}}
  \pgfpathlineto{\pgfpoint{0}{2.35\pgfutil@tempdima}}
  \pgfpathlineto{\pgfpoint{2\pgfutil@tempdima}{0pt}}
  \pgfpathlineto{\pgfpoint{0}{-2.35\pgfutil@tempdima}}
  \pgfpathlineto{\pgfpoint{0}{-\pgfutil@tempdima}}
  \pgfsetfillcolor{white}
  \pgfusepathqfillstroke
}
\makeatother

\newcommand{\fig}[1] {fig.~\ref{#1}}

\newcommand{\umlFont}[1]{\textsf{#1}}

\input{\PTPhDThesisDir/listing-styles.tex}

\usepackage{relsize}
\def\C++{%
    C\kern-.0667em\raise.25ex\hbox{\smaller[2]{++}}%
\spacefactor1000 }

\input{\PTPhDThesisDir/ecore-diagrams.tex}

%% file: listing-styles.tex
\usepackage{listings}
\lstdefinelanguage{ATL}{
  morekeywords={true,false,
  Bag,Set,OrderedSet,Sequence,Tuple,Integer,Real,Boolean,String,TupleType,
  not,and,or,xor,implies,module,create,from,uses,helper,def,context,
  rule,using,derived,to,mapsTo,distinct,
  foreach,in,do,if,then,else,endif,let,
  library,query,for,div,refining,entrypoint},
 keywordstyle=[2]{\textbf},
 morecomment=[l]{--},
 morestring=[b]{'},
 tabsize=1
}

\lstdefinelanguage{AcceleoC++}{
	morekeywords={auto,break,case,char,const,continue,default,do,double,%
		else,enum,extern,float,for,goto,if,int,long,register,return,%
		short,signed,sizeof,static,struct,switch,typedef,union,unsigned,%
		void,volatile,while},%
    sensitive    ,
    morecomment=[s]{/*}{*/},%
    morecomment=[s][\color{lightgray}]{[}{]},
    morecomment=[l]//,% nonstandard
	morestring=[b]",%
	morestring=[b]',%
	moredelim=*[directive]\#,%
	moredirectives={define,elif,else,endif,error,if,ifdef,ifndef,line,%
		include,pragma,undef,warning}%    
}[keywords,comments,strings,directives]

\lstdefinestyle{AcceleoTT}{
  basicstyle={\small\fontfamily{lmvtt}\selectfont},
  commentstyle={\small\fontfamily{lmvtt}\slshape},
  keywordstyle={\fontfamily{lmvtt}\fontseries{b}\selectfont}
}

\lstdefinestyle{black}{
  basicstyle={\small\fontfamily{lmtt}\selectfont},
  commentstyle={\small\fontfamily{ptm}\slshape},
  keywordstyle={\fontfamily{lmtt}\fontseries{b}\selectfont},
  columns=fullflexible,
  breaklines=true,
  mathescape=true,
  escapechar=\#,
  tabsize=4,
  frame=lines,
  showstringspaces=false,
  captionpos=b,
  float=tbp,
  deletecomment=[l]{--},
  moredelim=[l]{--},
  morekeywords={excludes,notEmpty,forAll,includes,subOrderedSet,indexOf,last,isEmpty,select,includesAll}
}

\lstdefinestyle{xmlblack}{
  basicstyle={\small\fontfamily{lmtt}\selectfont},
  commentstyle=\color{gray}\upshape,
  keywordstyle={\fontfamily{lmtt}\fontseries{b}\selectfont},
  columns=fullflexible,
  breaklines=true,
  mathescape=true,
  escapechar=\#,
  tabsize=4,
  frame=lines,
  showstringspaces=false,
  captionpos=b,
  float=tbp,
  morestring=[b]",
  morestring=[s]{>}{<},
  morecomment=[s]{<?}{?>},
  stringstyle=\color{black},
  identifierstyle=\color{darkblue},
  keywordstyle=\bf\ttfamily,
  morekeywords={xmlns,version,type}% list your attributes here
}

%% file: ecore-diagrams.tex
\usepackage{tikz}
\usetikzlibrary{arrows,calc,shapes.multipart,shadows}

\newcommand{\arity}[1]{
\mbox{
\begin{pgfscope}
\pgftext[top,y=1ex]{
\tikz{
\draw[] (-0.75ex,0) rectangle (0.75ex,0.75ex);
\node[rectangle,
align=center,
anchor=north,inner sep=1pt] {\sffamily\fontsize{6}{10}\selectfont #1};
}
}
\end{pgfscope}
}\rule[-2pt]{.25em}{0pt}%
}

\tikzstyle{umlfont} = [font=\sffamily]
\tikzstyle{inherit} = [-open triangle 45,umlfont]
\tikzstyle{compose} = [diamond-angle 60,umlfont]
\tikzstyle{refer} = [-angle 60,umlfont]
\tikzstyle{every class} = []
\tikzstyle{class} = [draw,umlfont,inner xsep=2pt,
rectangle split,
rectangle split parts=2, % split into 2 parts: NAME and MEMBERS.
rectangle split part align={center, left}, % Note: this does not work for minipage ('text width') envirionment
every text node part/.style={align=center},
align=left,
drop shadow,fill=white,
every class
]
\tikzstyle{atEnd} = [pos=0.7,anchor=south]
\tikzstyle{atStart} = [pos=0.3]

\newcommand{\isAbstract}[1] {\textsf{\textit{#1}}}

%% file: subsumption-meta.tikz
\begin{tikzpicture}[every text node part/.style={text centered}]
\node[class]
(System)
{
System
\nodepart{second}
\mbox{}\arity{1} name: EString \\
\mbox{}\arity{} description: EString
};

\node[class,above of=System,node distance=18ex]
(Module)
{
Module
\nodepart{second}
\mbox{}\arity{1} name: EString \\
\mbox{}\arity{} description: EString \\
\mbox{}\arity{1} layer: EInteger
%\mbox{}\arity{} period: EFloat
};

\node[class,right of=Module,node distance=15em,above=6ex]
(Input)
{
InputLine
\nodepart{second}
\mbox{}\arity{1} name: EString \\
\mbox{}\arity{} description: EString \\
\mbox{}\arity{1} dataType: EString
};

\node[class,right of=Module,node distance=15em,below=4ex]
(Output)
{
OutputLine
\nodepart{second}
\mbox{}\arity{1} name: EString \\
\mbox{}\arity{} description: EString \\
\mbox{}\arity{1} dataType: EString
};

\node[class,right of=Input,node distance=15em]
(Suppressor)
{
Suppressor
\nodepart{second}
\mbox{}\arity{1} time: EFloat\phantom{p}
};

\node[class,
%right of=Output,node distance=15em,below=6ex,
anchor=south
]
(Inhibitor) at (System.south -| Suppressor)
{
Inhibitor
\nodepart{second}
\mbox{}\arity{1} time: EFloat\phantom{p}
};

\node[class,right of=Output,node distance=25em,text width=7em]
(Modifier)
{
\isAbstract{Modifier}
\nodepart{second}
\mbox{}\phantom{p}
};

\draw[compose] (System) --
	%node [atEnd,left] {modules}
	node [atEnd,left] {0..*}
	(Module);

\draw[compose] ($(Module.north)!0.5!(Module.north east)$) |-
	%node [pos=0.85,above] {inputs}
	node [pos=0.85,below] {0..*}
	(Input);
\draw[compose] ($(Module.south)!0.5!(Module.south east)$) |-
	%node [pos=0.85,above] {outputs}
	node [pos=0.85,below] {0..*}
	(Output);
	
\draw[compose] (Input) --
	%node [pos=0.95,anchor=south east] {suppressedBy}
	node [pos=0.95,anchor=north east] {0..*}
	(Suppressor);

\draw[compose] (Output) |-
	%node [pos=0.95,anchor=south east] {inhibitedBy}
	node [pos=0.95,anchor=north east] {0..*}
	(Inhibitor);
	
\draw[refer,angle 60-angle 60] (Input) --
	node [pos=0.05,anchor=north east] {sink}
	node [pos=0.05,anchor=north west] {0..*}
	node [pos=0.95,anchor=south east] {source}
	node [pos=0.95,anchor=south west] {0..1}
	(Output);	

\draw[inherit] (Suppressor) -| (Modifier);
\draw[inherit] (Inhibitor) -| (Modifier);
\draw[refer,angle 60-angle 60] (Output) --
	node [pos=0.2,above] {controlledBy}
	node [pos=0.2,below] {1}
	node [pos=0.85,above] {controls\phantom{y}}
	node [pos=0.85,below] {0..*}
	(Modifier);
\end{tikzpicture}

%% file: submde.bbl
\begin{thebibliography}{10}
\providecommand{\url}[1]{#1}
\csname url@samestyle\endcsname
\providecommand{\newblock}{\relax}
\providecommand{\bibinfo}[2]{#2}
\providecommand{\BIBentrySTDinterwordspacing}{\spaceskip=0pt\relax}
\providecommand{\BIBentryALTinterwordstretchfactor}{4}
\providecommand{\BIBentryALTinterwordspacing}{\spaceskip=\fontdimen2\font plus
\BIBentryALTinterwordstretchfactor\fontdimen3\font minus
  \fontdimen4\font\relax}
\providecommand{\BIBforeignlanguage}[2]{{%
\expandafter\ifx\csname l@#1\endcsname\relax
\typeout{** WARNING: IEEEtran.bst: No hyphenation pattern has been}%
\typeout{** loaded for the language `#1'. Using the pattern for}%
\typeout{** the default language instead.}%
\else
\language=\csname l@#1\endcsname
\fi
#2}}
\providecommand{\BIBdecl}{\relax}
\BIBdecl

\bibitem{RSAaP:08}
D.~Kortenkamp and R.~Simmons, ``Robotic systems architectures and
  programming,'' in \emph{Springer Handbook of Robotics}, O.~Khatib and
  B.~Siciliano, Eds.\hskip 1em plus 0.5em minus 0.4em\relax Springer, 2008, pp.
  187--206.

\bibitem{brooks:1986_robust}
R.~Brooks, ``A robust layered control system for a mobile robot,''
  \emph{Robotics and Automation, IEEE Journal of}, vol.~2, no.~1, pp. 14--23,
  1986.

\bibitem{gat1998three}
E.~Gat, ``On three-layer architectures,'' in \emph{Artificial Intelligence and
  Mobile Robots}, D.~Kortenkamp, R.~P. Bonnasso, and R.~Murphy, Eds.\hskip 1em
  plus 0.5em minus 0.4em\relax AAAI Press, 1998, pp. 195--210.

\bibitem{Brugali:2009}
D.~Brugali and P.~Scandurra, ``{Component--based Robotic Engineering. Part I:
  Reusable building blocks},'' \emph{IEEE Robotics and Automation Magazine},
  vol.~16, no.~4, pp. 84--96, 2009.

\bibitem{schmidt2006model}
D.~Schmidt, ``Model-driven engineering,'' \emph{IEEE Computer}, vol.~39, no.~2,
  pp. 25--31, February 2006.

\bibitem{kleppe2009software}
A.~Kleppe, \emph{Software Language Engineering: Creating Domain-Specific
  Languages Using Metamodels}.\hskip 1em plus 0.5em minus 0.4em\relax
  Addison-Wesley, 2009.

\bibitem{kelly2008domain}
S.~Kelly and J.-P. Tolvanen, \emph{Domain-Specific Modeling: Enabling full code
  generation}.\hskip 1em plus 0.5em minus 0.4em\relax Wiley-IEEE Computer
  Society Press, April 2008.

\bibitem{msdsl2007}
S.~Cook, G.~Jones, S.~Kent, and A.~C. Wills, \emph{{Domain-Specific Development
  with Visual Studio DSL Tools}}.\hskip 1em plus 0.5em minus 0.4em\relax
  Addison-Wesley Professional, 2007.

\bibitem{emf2009}
D.~Steinberg, F.~Budinsky, M.~Paternostro, and E.~Merks, \emph{EMF: Eclipse
  Modeling Framework}, 2nd~ed.\hskip 1em plus 0.5em minus 0.4em\relax
  Addison-Wesley Professional, January 2009.

\bibitem{gronback2009eclipse}
R.~C. Gronback, \emph{{Eclipse Modeling Project: A Domain-Specific Language
  (DSL) Toolkit}}.\hskip 1em plus 0.5em minus 0.4em\relax Addison-Wesley
  Professional, 2009.

\bibitem{MOF}
\BIBentryALTinterwordspacing
{Object Management Group}, \emph{Meta Object Facility (MOF) Core Specification
  Version 2.0}, January 2006. [Online]. Available:
  \url{http://www.omg.org/spec/MOF/2.0/}
\BIBentrySTDinterwordspacing

\bibitem{OCL}
\BIBentryALTinterwordspacing
------, \emph{Object Constraint Language. Version 2.0}, May 2006. [Online].
  Available: \url{http://www.omg.org/spec/OCL/2.0/}
\BIBentrySTDinterwordspacing

\bibitem{MOFM2T:08}
\BIBentryALTinterwordspacing
------, \emph{MOF Model to Text Transformation Language}, January 2008.
  [Online]. Available: \url{http://www.omg.org/spec/MOFM2T/1.0/}
\BIBentrySTDinterwordspacing

\bibitem{warmer2003object}
J.~Warmer and A.~Kleppe, \emph{The Object Constraint Language: Getting Your
  Models Ready for MDA}.\hskip 1em plus 0.5em minus 0.4em\relax Addison-Wesley
  Longman Publishing Co., Inc. Boston, MA, USA, 2003.

\bibitem{umlSuper23}
\BIBentryALTinterwordspacing
{Object Management Group}, ``{OMG Unified Modeling Language (OMG UML),
  Superstructure, Version 2.3},'' OMG, Tech. Rep., May 2010. [Online].
  Available: \url{http://www.omg.org/spec/UML/2.3/Superstructure/PDF/}
\BIBentrySTDinterwordspacing

\bibitem{brooks1990behavior}
R.~Brooks, ``{The Behavior Language: User's Guide. A. I. Memo 1227.}''
  Cambridge, MA, USA, Tech. Rep., April 1990.

\bibitem{Thrun:2006Stanley}
S.~Thrun, M.~Montemerlo, H.~Dahlkamp, D.~Stavens, A.~Aron, J.~Diebel, P.~Fong,
  J.~Gale, M.~Halpenny, G.~Hoffmann, K.~Lau, C.~Oakley, M.~Palatucci, V.~Pratt,
  and P.~Stang, ``Stanley: The robot that won the {DARPA Grand Challenge},''
  \emph{Journal of Field Robotics}, vol.~23, no.~9, pp. 661--692, 2006.

\bibitem{POSIX-1.2008}
{IEEE and The Open Group}, \emph{Posix.1-2008}.\hskip 1em plus 0.5em minus
  0.4em\relax IEEE, 2008.

\bibitem{taft2006ada}
S.~T. Taft, R.~A. Duff, R.~L. Brukardt, E.~Ploedereder, and P.~Leroy, Eds.,
  \emph{Ada 2005 reference manual. Language and standard libraries.
  International standard ISO/IEC 8652/1995 (E) with technical corrigendum~1 and
  amendment~1}, ser. Lecture Notes in Computer Science.\hskip 1em plus 0.5em
  minus 0.4em\relax Springer, 2006, vol. 4348.

\bibitem{Gin:82}
G.~Gini and M.~Gini, ``{ADA}: {A}~language for robot programming?''
  \emph{Computers in Industry}, vol.~3, no.~4, pp. 253--259, 1982.

\bibitem{steele1994ada}
R.~D. Steele and P.~G. Backes, ``Ada and real-time robotics: lessons learned,''
  \emph{IEEE Computer}, vol.~27, no.~4, pp. 49--54, April 1994.

\bibitem{burns2004guide}
A.~Burns, B.~Dobbing, and T.~Vardanega, ``Guide for the use of the {A}da
  {R}avenscar {P}rofile in high integrity systems,'' \emph{ACM SIGAda Ada
  Letters}, vol.~24, no.~2, pp. 1--74, 2004.

\bibitem{vergnaud2004polyorb}
T.~Vergnaud, J.~Hugues, L.~Pautet, and F.~Kordon, ``{PolyORB}: {A}
  schizophrenic middleware to build versatile reliable distributed
  applications,'' in \emph{Reliable Software Technologies - Ada-Europe 2004},
  ser. Lecture Notes in Computer Science, A.~Llamosí and A.~Strohmeier,
  Eds.\hskip 1em plus 0.5em minus 0.4em\relax Springer Berlin Heidelberg, 2004,
  vol. 3063, pp. 106--119.

\bibitem{efftinge2006oaw}
S.~Efftinge and M.~V{\"o}lter, ``{oAW xText}: {A} framework for textual
  {DSLs},'' in \emph{Workshop on Modeling Symposium at Eclipse Summit},
  vol.~32, 2006.

\bibitem{HUTN}
\BIBentryALTinterwordspacing
{Object Management Group}, \emph{Human-Usable Textual Notation ({HUTN})
  Specification}, August 2004. [Online]. Available:
  \url{http://www.omg.org/spec/HUTN/1.0/}
\BIBentrySTDinterwordspacing

\bibitem{vaughan2008massively}
R.~Vaughan, ``Massively multi-robot simulation in stage,'' \emph{Swarm
  Intelligence}, vol.~2, no.~2, pp. 189--208, 2008.

\bibitem{fleury|iros97}
S.~Fleury, M.~Herrb, and R.~Chatila, ``\textsf{G$^{\mbox{en}}\!$oM}: {A} tool
  for the specification and the implementation of operating modules in a
  distributed robot architecture,'' \emph{Proceedings of the 1997 IEEE/RSJ
  International Conference on Intelligent Robots and Systems (IROS'97)},
  vol.~2, pp. 842--849, September 1997.

\bibitem{arias2010orccad}
S.~Arias, F.~Boudin, R.~Pissard-Gibollet, and D.~Simon, ``{ORCCAD}, robot
  controller model and its support using {E}clipse {M}odeling tools,'' in
  \emph{Proceedings of the 5th National Conference on Control Architectures of
  Robots CAR'10}, May 2010.

\bibitem{steck2010towards}
A.~Steck and C.~Schlegel, ``Towards quality of service and resource aware
  robotic systems through model-driven software development,'' in \emph{{1st
  International Workshop on Domain-Specific Languages and models for ROBotic
  systems (DSLRob'10)}}, 2010.

\end{thebibliography}
